\documentclass[11pt,letterpaper]{article}
\usepackage{naaclhlt2016}
\usepackage{times}
\usepackage{latexsym}
\usepackage{graphicx}
\usepackage{amssymb}
\graphicspath{ {images/} }
 
\naaclfinalcopy 
\def\naaclpaperid{***} 

\newcommand{\real}{\mathbb{R}}

\title{An Attentive Neural Architecture for Fine-grained Entity Type Classification}

\author{
    Sonse Shimaoka$^{\dagger}$\thanks{\hspace{0.55em}This work was conducted during a research visit to University College London.} \hspace{0.75em}
    Pontus Stenetorp$^{\ddagger}$ \hspace{0.75em}
    Kentaro Inui$^{\dagger}$ \hspace{0.75em}
    Sebastian Riedel$^{\ddagger}$ \\
    {\tt \{simaokasonse,inui\}@ecei.tohoku.ac.jp}\\
    {\tt \{p.stenetorp,s.riedel\}@cs.ucl.ac.uk}  \\
    $^{\dagger}$Graduate School of Information Sciences, Tohoku University \\
    $^{\ddagger}$Department of Computer Science, University College London
}
\date{}

\begin{document}

\maketitle

\begin{abstract}
In this work we propose a novel attention-based neural network model for the task of fine-grained entity type classification that unlike previously proposed models recursively composes representations of entity mention contexts.
Our model achieves state-of-the-art performance with $74.94\%$ loose micro F1-score on the well-established FIGER dataset, a relative improvement of $2.59\%$ .
We also investigate the behavior of the attention mechanism of our model and observe that it can learn contextual linguistic expressions that indicate the fine-grained category memberships of an entity.
\end{abstract}

\section{Introduction}
Entity type classification is the task of assigning semantic types to mentions of entities in sentences.
Identifying the types of entities is useful for various natural language processing tasks, such as relation extraction \cite{ling2012fine}, question answering \cite{lee2006fine}, and knowledge base population \cite{carlson2010coupled}.
Unfortunately, most entity type classification systems use a relatively small number of types (e.g. {\tt person}, {\tt organization}, {\tt location}, {\tt time}, and {\tt miscellaneous} \cite{grishman1996message}) which may be too coarse-grained for some NLP applications \cite{sekine2008extended}.
To address this shortcoming, a series of recent work has investigated entity type classification with a large set of fine-grained types \cite{lee2006fine,ling2012fine,yosef2012hyena,yogatama2015embedding,del2015finet}.

\begin{figure}[t]
    \includegraphics[width=8cm]{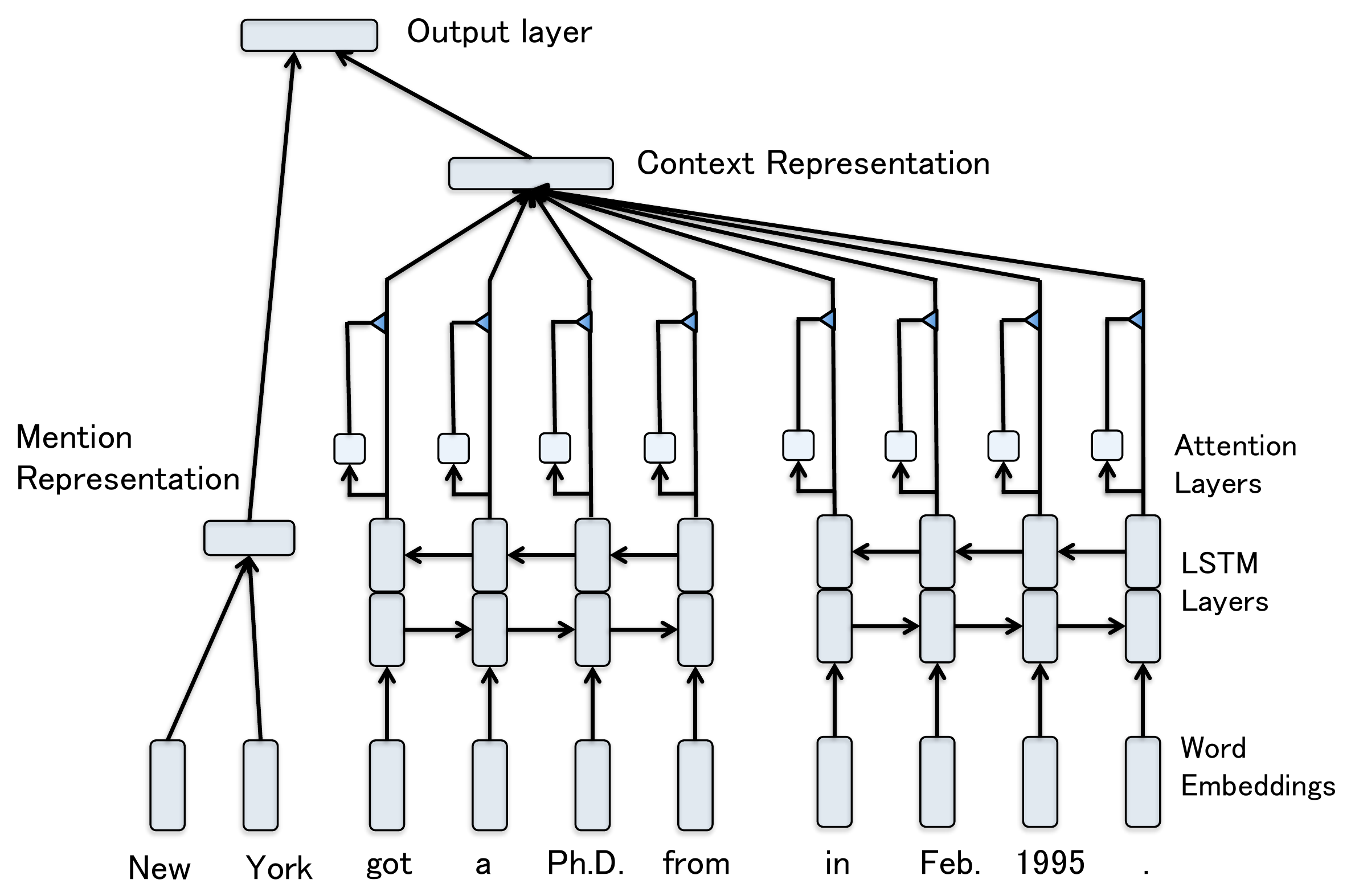}
    \centering
    \caption{An illustration of our proposed model predicting fine-grained semantic types for the mention ``New York'' in the sentence ``She got a Ph.D from New York in Feb. 1995.''.}
    \label{fig:model}
\end{figure}

Existing fine-grained entity type classification systems have used approaches ranging from sparse binary features to dense vector representations of entities to model the entity mention and its context.
However, no previously proposed system has attempted to learn to recursively compose representations of entity context.
For example, one can see that a phrase ``got a Ph.D. from'' is indicative of the next words being an educational institution, something which would be helpful for fine-grained entity type classification.

In this work our main contributions are two-fold:

\begin{enumerate}
    \item{A first model for fine-grained entity type classification that learns to recursively compose representations for the context of each mention and attains state-of-the-art performance on a well-established dataset.}
    \item{The observation that by incorporating an attention mechanism into our model, we not only achieve better performance, but also are able to observe that the model learns contextual linguistic expressions that indicate fine-grained category memberships of an entity.}
\end{enumerate}

\section{Related Work}

To the best of our knowledge, \newcite{lee2006fine} were the first to address the task of fine-grained entity type classification.
They defined 147 fine-grained entity types and evaluated a conditional random fields-based model on a manually annotated Korean dataset.
\newcite{sekine2008extended} advocated the necessity of a large set of types for entity type classification and defined $200$ types which served as a basis for future work on fine-grained entity type classification.

\newcite{ling2012fine} defined a set of $112$ types based on Freebase and created a training dataset from Wikipedia using a distant supervision method inspired by \newcite{mintz2009distant}.
For evaluation, they created a small manually annotated dataset of newspaper articles and also demonstrated that their system, FIGER, could improve the performance of a relation extraction system by providing fine-grained entity type predictions as features.
\newcite{yosef2012hyena} organised $505$ types in a hierarchical taxonomy, with several hundreds of types at different levels.
Based on this taxonomy they developed a multi-label hierarchical classification system.
In \newcite{yogatama2015embedding} the authors proposed to use label embeddings to allow information sharing between related labels.
This approach lead to improvements on the FIGER dataset, and they also demonstrated that fine-grained labels can be used as features to improve coarse-grained entity type classification performance.
\newcite{del2015finet} introduced the most fine-grained entity type classification system to-date, it operates on the the entire WordNet hierarchy with more than $16,000$ types.

While all previous models relied on hand-crafted features, \newcite{dong2015hybrid} defined $22$ types and created a two-part neural classifier.
They used a recurrent neural networks to recursively obtain a vector representation of each entity mention and used a fixed-size window to capture the context of each mention.
The key difference between our work and theirs lies in that we use recursive neural networks to compose context representations and that we employ an attention mechanism to allow our model to focus on relevant expressions.

\section{Models}

\subsection{Task Formulation}
We formulate the entity type classification problem as follows. 
Given an entity mention and its left and right context, our task is to predict its types.
Formally, the input is $l_{1},...,l_{C},m_{1},...,m_{M},r_1,...,r_C$, 
where $C$ is the window size of the left and right context, $l_{i}$ and $r_{i}$ represents a word in those contexts,  $M$ is the window size of the mention, and $m_i$ is a mention word.
If a context or a mention extends beyond the sentence length, a padding symbol is used in-place of a word.
Given this input we compute a probability $y_{k} \in \real$ for each of the $K$ types.

At inference, the type $k$ is predicted if $y_{k}$ is greater than $0.5$ or $y_{k}$ is the maximum value $\forall k \in K$.
The motivation of the former is that it acts as a cut-off, while the latter enforces the constraint that each mention is assigned at least one type.

\subsection{General Model}
While both mentions and contexts play important roles in determining the types, the complexity of learning to represent them are different.
During initial experiments, we observed that our model could learn from mentions significantly easier than from the context, leading to poor model generalization.
This motivated us to use different models for modeling mentions and contexts.
Specifically, all of our models described below firstly compute a mention representation $v_m \in \real^{D_m \times 1}$ and context representation $v_c \in \real^{D_c \times 1}$ separately, and then concatenate them to be passed to the final logistic regression layer with weight matrix $W_y \in \real^{ K \times (D_m+D_c) }$:
\begin{equation}
    y = \frac{1}{1 + \exp \left( -W_y \left[ \begin{array}{c} v_{m} \\ v_{c}\\ \end{array} \right] \right)}
\end{equation}

Note that we did not include a bias term in the above formulation since the type distribution in the training and test corpus could potentially be significantly different due to domain differences.
That is, in logistic regression, a bias fits to the empirical distribution of types in the training set, which would lead to bad performance on a test set that has a different type distribution.

The loss $L$ for a prediction $y$ when the true labels are encoded in a binary vector $t \in \{0,1\}^{K \times 1}$ is the following cross entropy loss function:
\begin{equation}
    L(y,t) = \sum_{k=1}^K - t_{k}\log(y_{k}) - (1 - t_{k})\log(1 - y_{k}) 
\end{equation}

\subsection{Mention Representation}
Mention representations are computed by averaging all the embeddings of the words in the mention.
Let the vocabulary be $V$ and the function $u : V \mapsto \real^{D_m \times 1}$ be a mapping from a word to its embedding.
Formally, the mention representation $v_m$ is obtained as follows.
\begin{equation}
   v_{m} = \frac{1}{M} \sum_{i=1}^{M} u(m_{i})
\end{equation}

During our experiments we were surprised by the fact that unlike the observations made by \newcite{dong2015hybrid}, complex neural models did not work well for learning mention representations compared to the simpler model described above.
One possible explanation for this would be labeling discrepancies between the training and test set.
For example, the label {\tt time} is assigned to days of the week (e.g. ``Friday'', ``Monday'', and ``Sunday'') in the test set, but not in the training set, whereas explicit dates (e.g. ``Feb. 24'' and ``June 4th'') are assigned the {\tt time} label in both the training and test set.
This may be harmful for complex models due to their tendency to overfit on the training data.

\subsection{Context Representation}
We compare three methods for computing context representations. 
\subsubsection{Averaging Encoder}
Applying the same averaging approach as for the mention representation for both the left and right context.
Thus, the concatenation of those two vectors becomes the representation of the context:
\begin{eqnarray}
    v_c =  \frac{1}{C} \sum_{i=1}^{C}　\left[ \begin{array}{c}   u(l_{i}) \\ u(r_{i}) \\ \end{array} \right] 
\end{eqnarray}

\subsubsection{LSTM Encoder}
The left and right context are encoded recursively using an LSTM cell \cite{hochreiter1997long}.
Given an input embedding $u_i \in \real^{D_m \times 1}$, the previous output $h_{i-1} \in \real^{D_h \times 1}$, and the previous cell state $s_{i-1} \in \real^{D_h \times 1}$, the high-level formulation of the recursive computation by an LSTM cell is as follows:
\begin{equation}
    h_i, s_i = lstm(u_i,h_{i-1},s_{i-1})
\end{equation}

For the left context, the model reads sequences $ l_{1},...,l_{C} $ from left to right to produce the outputs
$\overrightarrow{h_{1}^l},..., \overrightarrow{h_{C}^l}$.
For the right context, the model reads sequences $ r_{C},...,r_{1} $ from right to left to produce the outputs
$\overleftarrow{h_{1}^r},..., \overleftarrow{h_{C}^r}$.
Then the representation $v_c$ is obtained by concatenating  $\overrightarrow{h_{C}^l}$ and $\overleftarrow{h_{1}^r}$:
\begin{equation}
    v_c = \left[ \begin{array}{c}  \overrightarrow{h_{C}^l} \\ \overleftarrow{h_{1}^r} \\ \end{array} \right] 
\end{equation}
A more detailed formulation of the LSTM used in this work can be found in \newcite{sak2014long}.

\subsubsection{Attentive Encoder}
While an LSTM can encode sequential data, it still finds it difficult to learn long-term dependencies.
Inspired by recent work using attention mechanisms for natural language processing \cite{hermann2015teaching,rocktaschel2015reasoning}, we circumvent this problem by introducing a novel attention mechanism.
We also hypothesize that by incorporating an attention mechanism the model can recognize informative expressions for the classification and make the model behavior more interpretable.

The computation of the attention mechanism is as follows. 
Firstly, for both the right and left context, we encode the sequences using bi-directional LSTMs \cite{graves2012supervised}.
We denote the outputs as $\overrightarrow{h_{1}^l},\overleftarrow{h_{1}^l},..., \overrightarrow{h_{C}^l},\overleftarrow{h_{C}^l}$ 
and $\overrightarrow{h_{1}^r},\overleftarrow{h_{1}^r},..., \overrightarrow{h_{C}^r},\overleftarrow{h_{C}^r}$. 

For each output layer of the bi-directional LSTMs, we compute a scalar value $\tilde{a}_{i} \in \real$ using a two-layer feed forward neural network $e_i \in \real^{D_a \times 1}$ and weight matrices $W_e \in \real^{ D_a \times 2D_h }$ and $W_a \in \real^{ 1 \times D_a}$.
We then normalize these scalar values such that they sum to $1$.
We refer to these normalized scalar values $a_i \in \real$ as attentions.
Lastly, we take a weighted sum of the output layers of the bidirectional LSTMs as the representation of the context weighted by the attentions $a_i$:
\begin{eqnarray}
    e_{i}^l &=& \tanh \left( W_e \left[ \begin{array}{c} \overrightarrow{h_i^l} \\ \overleftarrow{h_i^l} \\ \end{array} \right] \right) \\
    \tilde{a}_{i}^l &=& \exp(W_a e_i^l) \\
    a_{i}^l &=& \frac{\tilde{a}_{i}^l}{\sum_{i=1}^C \tilde{a}_{i}^l + \tilde{a}_{i}^r} \\
    v_{c} &=& \sum_{i=1}^{C} a_{i}^l  \Biggl[ \begin{array}{c} \overrightarrow{h_i^l} \\ \overleftarrow{h_i^l} \\ \end{array} \Biggr] 
           + a_{i}^r  \Biggl[ \begin{array}{c} \overrightarrow{h_i^r} \\ \overleftarrow{h_i^r} \\ \end{array} \Biggr]
\end{eqnarray}
The equations for computing $e_{i}^r$, $\tilde{a}_{i}^r$, and $a_{i}^r$ were omitted for brevity and the overall picture of our proposed model is illustrated in Figure \ref{fig:model}.
\section{Experiment}

\begin{figure*}[t]
    \includegraphics[width=16cm]{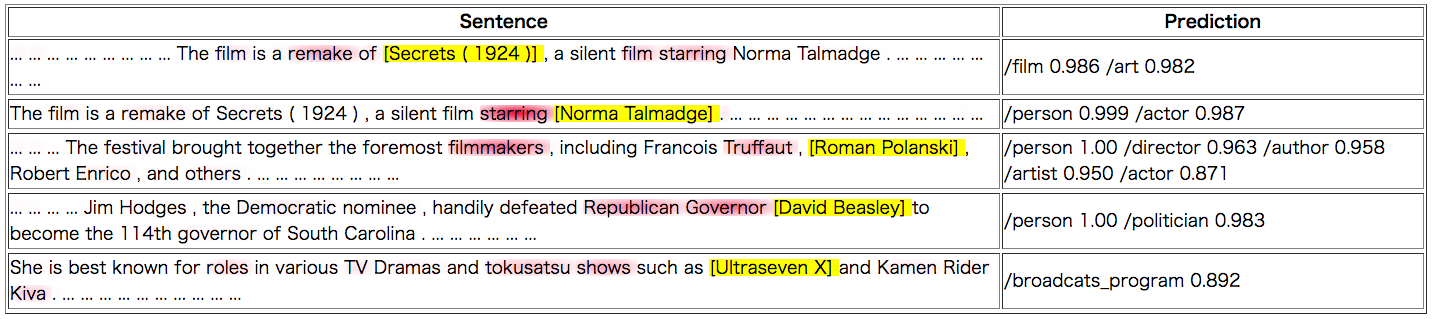}
    \centering
    \caption{Examples of our model attending over contexts for a given mention.}
    \label{fig:examples}
\end{figure*}

\subsection{Dataset}
To train and evaluate our model we use the publicly available FIGER dataset with 112 fine-grained types from \newcite{ling2012fine}.
The sizes of our datasets are $2,600,000$ for training, $90,000$ for development, and $563$ for testing.
Note that the train and development sets were created from Wikipedia, whereas the test set is a manually annotated dataset of newspaper articles.

\subsection{Pre-trained Word Embeddings}
The only features used by our model are pre-trained word embeddings that were not updated during training to help the model generalize for words not appearing in the training set.
Specifically, we used the freely available $300$ dimensional cased word embeddings trained on 840 billion tokens from the Common Crawl supplied by \newcite{pennington2014glove}.
As embeddings for out-of-vocabulary words, we used the embedding of the ``unk'' token from the pre-trained embeddings.

\subsection{Evaluation Criteria}
Following \newcite{ling2012fine}, we evaluate the model performances by strict, loose macro, and loose micro measures.
For the $i$-th instance, let the set of the predicted types be $\hat{T}_i$, and the set of the true types be $T_i$.
Then the precisions and recall for each measure are computed as follows.
\begin{itemize}
    \item strict \begin{equation} Precision = Recall = \frac{1}{N} \sum_{i=1}^N \delta (\hat{T}_i=T_i)  \end{equation} 
    \item loose macro \begin{eqnarray} Precision = \frac{1}{N} \sum_{i=1}^N \frac{|\hat{T}_i \cap T_i|}{|\hat{T}_i|} \\
                                       Recall    =  \frac{1}{N} \sum_{i=1}^N \frac{|\hat{T}_i \cap T_i|}{|T_i|}
                      \end{eqnarray} 
    \item loose micro \begin{eqnarray} Precision = \frac{ \sum_{i=1}^N  |\hat{T}_i \cap T_i| }{ \sum_{i=1}^N |\hat{T}_i|} \\
                                       Recall    = \frac{ \sum_{i=1}^N |\hat{T}_i \cap T_i| }{ \sum_{i=1}^N |T_i|}
                        \end{eqnarray} 
\end{itemize}
Where $N$ is the total number of instances.
\subsection{Hyperparameter Settings}
As hyperparameters, all three models used the same $D_m = 300$ dimensional word embeddings, the hidden-size of the LSTM was set to $D_h = 100$, and the hidden-layer size of the attention module was set to $D_a = 50$.
We used Adam \cite{kingma2014adam} as our optimization method with a learning rate of $0.005$ with a mini-batch size of $1,000$.
As a regularizer we used dropout with probability $0.5$ applied to the mention representation.

The context window size was set to $C=15$ and mention window size was set to $M=5$.
It should be noted that our approach is not restricted to using fixed window sizes, rather this is an implementation detail arising from current limitations of the machine learning library used when handling dynamic-width recurrent neural networks.
For each epoch we iterated over the training data set ten times and then evaluated the model performance on the development set.
After training we picked up the best model on the development set as our final model and report the performance on the test set.
Our model implementation was done in Python using the TensorFlow \cite{abadi2015tensorflow} machine learning library.

\subsection{Results}

The performance of the various models are summarized Tables~\ref{results} and~\ref{results2}.
We see that the Averaging base line performs well in spite of its relative simplicity, the LSTM model shows some improvements, and the attention model performs better than any previously proposed method.
In Figure~\ref{fig:examples}, we visualize the attentions for several instances that were manually selected from the development set.
It is clear that our proposed model is attending over expressions relevant for the entity types such as immediately adjacent to the mention such as ``starring'' and ``Republican Governor'', as well as more distant expressions such as ``filmmakers''.

\begin{table}[t]
\centering
\begin{tabular}{|l|r|r|r|}
    \hline
    Models                                 & P     & R     & F1         \\ \hline
    \newcite{ling2012fine}                 & -     & -     & 69.30      \\ \hline
    \newcite{yogatama2015embedding}        & \bf{82.23} & 64.55 & 72.35      \\ \hline
    Averaging Encoder                      & 68.63 & 69.07 & 68.65      \\ \hline
    LSTM Encoder                           & 72.32 & 70.36 & 71.34      \\ \hline
    Attentive Encoder                      & 73.63 & \bf{76.29} & \bf{74.94} \\ \hline
\end{tabular}
    \caption{Loose Micro Precision (P), Recall (R), and F1-score on the test set}
    \label{results}
\end{table}

\begin{table}[t]
\centering
\begin{tabular}{|l|r|r|r|}
    \hline
    Models                          & Strict        & \shortstack{Loose \\ Macro}   & \shortstack{Loose \\ Micro}   \\ \hline
    \newcite{ling2012fine}          & 52.30         & 69.90                         & 69.30       \\ \hline
    \newcite{yogatama2015embedding} & -             & -                             & 72.25       \\ \hline
    Averaging Encoder               & 51.89         & 72.24                         & 68.65     \\ \hline
    LSTM Encoder                    & 55.60         & 73.95                         & 71.34    \\ \hline
    Attentive Encoder               & \bf{58.97}    & \bf{77.96}                    & \bf{74.94}       \\ \hline
\end{tabular}
    \caption{Strict, Loose Macro and Loose  Micro F1-scores}
    \label{results2}
\end{table}

\section{Conclusion}
In this paper, we proposed a novel state-of-the-art neural network architecture with an attention mechanism for the task of fine-grained entity type classification.
We also demonstrated that the model can successfully learn to attend over expressions that are important for the classification of fine-grained types.

\section*{Acknowledgments}
This work was supported by CREST-JST, JSPS KAKENHI Grant Number 15H01702, a Marie Curie Career Integration Award, and an Allen Distinguished Investigator Award.
We would like to thank the anonymous reviewers and Koji Matsuda for their helpful comments and feedback.

\bibliography{naaclhlt2016}

\begin{thebibliography}{}

\bibitem[\protect\citename{Abadi \bgroup et al.\egroup
  }2015]{abadi2015tensorflow}
Mart\'{\i}n Abadi, Ashish Agarwal, Paul Barham, Eugene Brevdo, Zhifeng Chen,
  Craig Citro, Greg~S. Corrado, Andy Davis, Jeffrey Dean, Matthieu Devin,
  Sanjay Ghemawat, Ian Goodfellow, Andrew Harp, Geoffrey Irving, Michael Isard,
  Yangqing Jia, Rafal Jozefowicz, Lukasz Kaiser, Manjunath Kudlur, Josh
  Levenberg, Dan Man\'{e}, Rajat Monga, Sherry Moore, Derek Murray, Chris Olah,
  Mike Schuster, Jonathon Shlens, Benoit Steiner, Ilya Sutskever, Kunal Talwar,
  Paul Tucker, Vincent Vanhoucke, Vijay Vasudevan, Fernanda Vi\'{e}gas, Oriol
  Vinyals, Pete Warden, Martin Wattenberg, Martin Wicke, Yuan Yu, and Xiaoqiang
  Zheng.
\newblock 2015.
\newblock {TensorFlow}: Large-scale machine learning on heterogeneous systems.

\bibitem[\protect\citename{Carlson \bgroup et al.\egroup
  }2010]{carlson2010coupled}
Andrew Carlson, Justin Betteridge, Richard~C Wang, Estevam~R Hruschka~Jr, and
  Tom~M Mitchell.
\newblock 2010.
\newblock Coupled semi-supervised learning for information extraction.
\newblock In {\em Proceedings of the third ACM international conference on Web
  search and data mining}, pages 101--110. ACM.

\bibitem[\protect\citename{Del~Corro \bgroup et al.\egroup }2015]{del2015finet}
Luciano Del~Corro, Abdalghani Abujabal, Rainer Gemulla, and Gerhard Weikum.
\newblock 2015.
\newblock Finet: Context-aware fine-grained named entity typing.
\newblock In {\em Conference on Empirical Methods in Natural Language
  Processing}, pages 868--878. ACL.

\bibitem[\protect\citename{Dong \bgroup et al.\egroup }2015]{dong2015hybrid}
Li~Dong, Furu Wei, Hong Sun, Ming Zhou, and Ke~Xu.
\newblock 2015.
\newblock A hybrid neural model for type classification of entity mentions.
\newblock In {\em Proceedings of the 24th International Conference on
  Artificial Intelligence}, pages 1243--1249. AAAI Press.

\bibitem[\protect\citename{Graves}2012]{graves2012supervised}
Alex Graves.
\newblock 2012.
\newblock {\em Supervised sequence labelling}.
\newblock Springer.

\bibitem[\protect\citename{Grishman and Sundheim}1996]{grishman1996message}
Ralph Grishman and Beth Sundheim.
\newblock 1996.
\newblock Message understanding conference-6: A brief history.
\newblock In {\em COLING}, volume~96, pages 466--471.

\bibitem[\protect\citename{Hermann \bgroup et al.\egroup
  }2015]{hermann2015teaching}
Karl~Moritz Hermann, Tomas Kocisky, Edward Grefenstette, Lasse Espeholt, Will
  Kay, Mustafa Suleyman, and Phil Blunsom.
\newblock 2015.
\newblock Teaching machines to read and comprehend.
\newblock In {\em Advances in Neural Information Processing Systems}, pages
  1684--1692.

\bibitem[\protect\citename{Hochreiter and Schmidhuber}1997]{hochreiter1997long}
Sepp Hochreiter and J{\"u}rgen Schmidhuber.
\newblock 1997.
\newblock Long short-term memory.
\newblock {\em Neural computation}, 9(8):1735--1780.

\bibitem[\protect\citename{Kingma and Ba}2014]{kingma2014adam}
Diederik Kingma and Jimmy Ba.
\newblock 2014.
\newblock Adam: A method for stochastic optimization.
\newblock {\em arXiv preprint arXiv:1412.6980}.

\bibitem[\protect\citename{Lee \bgroup et al.\egroup }2006]{lee2006fine}
Changki Lee, Yi-Gyu Hwang, Hyo-Jung Oh, Soojong Lim, Jeong Heo, Chung-Hee Lee,
  Hyeon-Jin Kim, Ji-Hyun Wang, and Myung-Gil Jang.
\newblock 2006.
\newblock Fine-grained named entity recognition using conditional random fields
  for question answering.
\newblock In {\em Information Retrieval Technology}, pages 581--587. Springer.

\bibitem[\protect\citename{Ling and Weld}2012]{ling2012fine}
Xiao Ling and Daniel~S Weld.
\newblock 2012.
\newblock Fine-grained entity recognition.
\newblock In {\em In Proc. of the 26th AAAI Conference on Artificial
  Intelligence}. Citeseer.

\bibitem[\protect\citename{Mintz \bgroup et al.\egroup }2009]{mintz2009distant}
Mike Mintz, Steven Bills, Rion Snow, and Dan Jurafsky.
\newblock 2009.
\newblock Distant supervision for relation extraction without labeled data.
\newblock In {\em Proceedings of the Joint Conference of the 47th Annual
  Meeting of the ACL and the 4th International Joint Conference on Natural
  Language Processing of the AFNLP: Volume 2-Volume 2}, pages 1003--1011.
  Association for Computational Linguistics.

\bibitem[\protect\citename{Pennington \bgroup et al.\egroup
  }2014]{pennington2014glove}
Jeffrey Pennington, Richard Socher, and Christopher~D Manning.
\newblock 2014.
\newblock Glove: Global vectors for word representation.
\newblock In {\em EMNLP}, volume~14, pages 1532--1543.

\bibitem[\protect\citename{Rockt{\"a}schel \bgroup et al.\egroup
  }2015]{rocktaschel2015reasoning}
Tim Rockt{\"a}schel, Edward Grefenstette, Karl~Moritz Hermann, Tom{\'a}{\v{s}}
  Ko{\v{c}}isk{\`y}, and Phil Blunsom.
\newblock 2015.
\newblock Reasoning about entailment with neural attention.
\newblock {\em arXiv preprint arXiv:1509.06664}.

\bibitem[\protect\citename{Sak \bgroup et al.\egroup }2014]{sak2014long}
Hasim Sak, Andrew~W Senior, and Fran{\c{c}}oise Beaufays.
\newblock 2014.
\newblock Long short-term memory recurrent neural network architectures for
  large scale acoustic modeling.
\newblock In {\em INTERSPEECH}, pages 338--342.

\bibitem[\protect\citename{Sekine}2008]{sekine2008extended}
Satoshi Sekine.
\newblock 2008.
\newblock Extended named entity ontology with attribute information.
\newblock In {\em LREC}, pages 52--57.

\bibitem[\protect\citename{Yogatama \bgroup et al.\egroup
  }2015]{yogatama2015embedding}
Dani Yogatama, Dan Gillick, and Nevena Lazic.
\newblock 2015.
\newblock Embedding methods for fine grained entity type classification.
\newblock In {\em Proceedings of the 53rd Annual Meeting of the Association for
  Computational Linguistics and the 7th International Joint Conference on
  Natural Language Processing of the Asian Federation of Natural Language
  Processing, ACL}, pages 26--31.

\bibitem[\protect\citename{Yosef \bgroup et al.\egroup }2012]{yosef2012hyena}
Mohamed~Amir Yosef, Sandro Bauer, Johannes Hoffart, Marc Spaniol, and Gerhard
  Weikum.
\newblock 2012.
\newblock Hyena: Hierarchical type classification for entity names.
\newblock In {\em 24th International Conference on Computational Linguistics},
  pages 1361--1370. ACL.

\end{thebibliography}
\bibliographystyle{naaclhlt2016}

\end{document}